\documentclass[sigconf]{acmart}
\usepackage[utf8]{inputenc} 
\usepackage[T1]{fontenc}    
\usepackage{hyperref}       
\usepackage{url}            
\usepackage{booktabs}       
\usepackage{amsfonts}       
\usepackage{nicefrac}       
\usepackage{microtype}      
\usepackage{lipsum}		
\usepackage{graphicx}
\usepackage{natbib}
\usepackage{doi}
\usepackage{booktabs} 
\usepackage{subcaption}
\usepackage{graphicx}
\usepackage{diagbox}
\usepackage{tikz}

\usepackage{algorithm}
\usepackage{algpseudocode}
\usepackage{xcolor} 
\usepackage{tikz}
\usetikzlibrary{shapes.geometric, arrows, positioning, automata,positioning,fit,backgrounds}
\tikzstyle{process} = [rectangle, minimum width=2cm, minimum height=1cm, text centered, draw=black, fill=white!30]
\tikzstyle{sum} = \tikzstyle{sum} = [draw, circle, minimum size=.5cm]
\tikzstyle{arrow} = [thick,->,>=stealth]

\usepackage{amsmath}

\usepackage{bm}
\usepackage{units}
\usepackage{float}
\usepackage{dblfloatfix}
\usepackage{algorithm}

\AtBeginDocument{%
  \providecommand\BibTeX{{%
    \normalfont B\kern-0.5em{\scshape i\kern-0.25em b}\kern-0.8em\TeX}}}

\setcopyright{acmcopyright}
\copyrightyear{2022}
\acmYear{2022}
\acmDOI{10.1145/1122445.1122456}

\acmConference[GECCO '22]{The Genetic and Evolutionary Computation Conference 2022}{July 9--13, 2022}{Boston, USA}
\acmPrice{15.00}
\acmISBN{978-1-4503-XXXX-X/18/06}

\begin{document}

\title{Environment induced emergence of collective behaviour in evolving swarms with limited sensing}

\author{Fuda van Diggelen}
\authornote{Three authors contributed equally to this research.}
\email{fuda.van.diggelen@vu.nl}
\author{Jie Luo}
\authornotemark[1]
\author{Tugay Alperen Karag{\"u}zel}
\authornotemark[1]
\affiliation{%
  \institution{Vrije Universiteit Amsterdam}
  \streetaddress{De Boelaan 1111}
  \state{Noord-Holland} 
  \postcode{1081 HV}
  \country{The Netherlands}
}

\author{Nicolas Cambier}
\affiliation{%
  \institution{Vrije Universiteit Amsterdam}
  \streetaddress{De Boelaan 1111}
  \state{Noord-Holland} 
  \postcode{1081 HV}
  \country{The Netherlands}
}
\email{n.p.a.cambier@vu.nl}

\author{Eliseo Ferrante}
\affiliation{%
  \institution{Vrije Universiteit Amsterdam}
  \streetaddress{De Boelaan 1111}
  \state{Noord-Holland} 
  \country{The Netherlands}
}
\affiliation{%
  \institution{Technology Innovation Institute}
  \city{Abu Dhabi}
  \country{United Arab Emirates}
}
\email{e.ferrante@vu.nl}

\author{A.E. Eiben}
\affiliation{%
  \institution{Vrije Universiteit Amsterdam}
  \streetaddress{De Boelaan 1111}
  \state{Noord-Holland} 
  \postcode{1081 HV}
  \country{The Netherlands}
}
\email{a.e.eiben@vu.nl}


\begin{abstract}
Designing controllers for robot swarms is challenging, because human developers have typically no good understanding of the link between the details of a controller that governs individual robots and the swarm behavior that is an indirect result of the interactions between swarm members and the environment. In this paper we investigate whether an evolutionary approach can mitigate this problem. We consider a very challenging task where robots with limited sensing and communication abilities must follow the gradient of an environmental feature and use Differential Evolution to evolve a neural network controller for simulated robots. We conduct a systematic study to measure the flexibility and scalability of the method by varying the size of the arena and number of robots in the swarm. The experiments confirm the feasibility of our approach, the evolved robot controllers induced swarm behavior that solved the task. We found that solutions evolved under the harshest conditions (where the environmental clues were the weakest) were the most flexible and that there is a sweet spot regarding the swarm size. Furthermore, we observed collective motion of the swarm, showcasing truly emergent behavior that was not represented in- and selected for during evolution.

\end{abstract}




\maketitle

\section{Introduction}
Animal groups in nature have evolved collective motion behaviors for certain benefits, like increasing environmental awareness \cite{couzin2005-animalgroups, kearns2010-bacterialswarming} and safety from predation \cite{olson2013-predatorconfusion}. Implementing such behaviors in robot swarms has several applications, but designing adequate robot controllers is a great challenge. The principal problem is that the controller can only govern the individual robots directly, but the desired behavior is defined at the group level \cite{hasselmann2021empirical}. Human developers have typically no good understanding of the link between the controller details and the induced swarm behavior that is an indirect result of the interactions between individual swarm members and the environment. In the current practice, controllers in swarm robotics require an extensive manual design and fine-tuning process before obtaining desired behavior. 

We developed an artificial evolutionary pipeline to automatically design and evaluate a robot controller for the swarm to complete a collective level task. On a conceptual level, we may note that we hereby address a higher level of complexity than traditional EC, even one step higher than evolutionary robotics (ER). As explained in \cite{Eiben2015Introduction-to}, in traditional EC we have a 3-step chain from genotypes to fitness values, genotype $\rightarrow$ phenotype $\rightarrow$ fitness, while in ER the chain is 4-fold, genotype $\rightarrow$ phenotype $\rightarrow$ behavior $\rightarrow$ fitness. In the case of robot swarms, we have one more level, that of the group behavior: genotype $\rightarrow$ phenotype $\rightarrow$ individual behavior $\rightarrow$ group behavior $\rightarrow$ fitness. Obviously, the arrow from individual behavior to group behavior is complex in itself, thus optimizing along the whole chain from genotype to fitness is very intricate. This makes the endeavour of evolving controllers for swarms far from trivial. 

For this paper we consider a very challenging task, where a swarm consisting of robots with very limited sensing and communication abilities must follow the gradient of an environmental feature, e.g., light intensity, temperature, or radiation. To solve this problem, we define genotypes that represent neural networks (phenotypes) in a robot and postulate that all robots of the swarm have the same controller. The fitness of a genotype / phenotype will be defined by the behavior of the swarm. Our goal is to investigate whether an evolutionary algorithm is capable of finding controllers that enable the swarm to follow the gradient. 

By constraining a single agent's sensor capabilities to only sense the local scalar value of the gradient, we know that this task is not solvable for an agent in isolation without any memory to compute the gradient on its own. This problem is especially interesting as we thus investigate if an environment can induce a collective behavior which would normally not evolve, to complete a task that otherwise could not be solved. In contrast to other works, the optimization does not directly solve for emergent behavior but might exhibit it as a successful strategy.
To this end, we are seeking answers to the following specific research questions.

{\textbf{Research Question 1:}}
Will there be any emergent collective behavior among the members of the swarm? 

{\textbf{Research Question 2:}}
How flexible and scalable are the evolved solutions with respect to changes in the size of the robot arena and the number of robots in the swarm?

To answer our research questions, we set up a system where the controllers of (simulated) swarms of differential drive robots are evolved by Differential Evolution. We measure the flexibility and scalability by running experiments in three arenas and with three different group sizes in a simulator called Isaac Gym \cite{makoviychuk2021isaac}. 




\section{Related Work}
Automated design for (swarm) robotics has a long history with varying approaches \cite{ prabhu2018survey, birattari2021automode}, which often require an optimization method that can handle high non-linearity and non-smooth objective functions. EAs proved themselves to be capable of handling such a hard task, which resulted in the development of fields like evolutionary robotics \cite{floreano2000evolutionary}. In this paper, we draw inspiration from this field and create an automated design pipeline suitable for swarm robotics. A strong difference in our approach is the level of control that we specify. Evolutionary robotics is often focused on learning low-level controls on a single robot for basic tasks like locomotion \cite{nelson2009fitness, Wright2015}. In contrast, a swarm controller indirectly specifies a higher level of behavior on the group level \cite{trianni2014evolutionary}. This additional level of abstraction creates a situation where we cannot expect a group behaviour to directly correlate with the objective. Nevertheless as mentioned before, the ability to optimize control when there is no clear mapping from variables to objective is what makes evolution attractive to apply on swarms as well. 


Indeed, non-automated designs usually struggle to resolve this mapping between agent-based (i.e. microscopic) and group-level (i.e. macroscopic) models, and must therefore resort to trial-and-error. In some cases, it is however possible to establish a quantitative micro-macro link so that mathematical equations are able to indicate the settings required to achieve an objective or, vice versa, to predict the outcome of an experiment from variables' values \cite{reina2015quantitative}. Such a micro-macro link is possible for binary (leftward or rightward) alignment of multiple agents \cite{hamann2014derivation}, but it is inapplicable to more advanced alignment-dependent behaviors such as flocking \cite{reynolds1987flocks}. As a matter of fact, existing approaches are only relevant to a fairly limited set of tasks; specifically collective decision-making between a finite quantity of options, i.e., best-of-\textit{n} problems \cite{valentini2017best}.

Typically, collective perception can be represented as a best-of-\textit{n} problem \cite{Valentini2016-Collectivefeature}, but continuous representations are more susceptible to be found in real environments \cite{puckett2018collective,Berdahl2018-animalnavigation,Wahby2019-swarmdensity}. 
In formal definition collective perception is; \textit{where social interactions among individuals lead to collective computation of an environmental property by only allowing scalar measurements made by these individuals} \cite{Berdahl2018-animalnavigation}. This approach has been proven viable in biology \cite{puckett2018collective, berdahl2013emergent}, where schools of fish --incapable of sensing the light gradient-- can nevertheless follow the gradient to hide in the dark, by only considering their interactions with other fish. 

Collective perception is also studied on artificial systems. In \cite{Wahby2019-swarmdensity}, a honey-bee inspired algorithm was used for aggregation of a robot swarm as a result of local interactions between each other and an environmental feature. Unfortunately, the authors provided a discrete environmental feature to the robots plus some additional (limited) communication capabilities, which made the resulting swarm behavior not truly obtained through a collective estimation alone. Another collective perception application can be found in \cite{Valentini2016-Collectivefeature} where robots collectively estimate certain environmental features and decide about the most frequent one (i.e. white tiles or black tiles in a tile grid). Yet again collective perception was not strictly necessary as the agents in \cite{Valentini2016-Collectivefeature} could communicate through voting and direct information exchange. There are many more papers demonstrating collective behavior of robot swarms where some form of collectively perceived feature(s) is used. Unfortunately, most of them either use communication capabilities or directional information related to the environment and/or other agents \cite{Schmickl2016-FSTaxis, Varughese2019-swarmfstaxis}. An exception to the rule is presented by \cite{karaguzel2020collective}. Here, a true collective perception method is proposed to control a flocking robot swarm that can sense a gradient, in an emergent way, and follow it in the environment, using only local and scalar measurements. The only downside to this solution is the manual optimization and design of the governing rules, which is a slow process that requires a lot of knowledge. Additionally, such an approach is not easily generalizable among other tasks and can easily be affected from individual failures or varying environmental conditions.

To cope with setups that are not fully specified at design time, automated design has been applied for collective perception tasks as well. \cite{Campo2011-selforg-discrim} employs an evolutionary optimization framework to optimize controller parameters of robots in a swarm, to learn to discriminate different shelters in the environment and aggregate at the best to exploit. Although successful, \cite{Campo2011-selforg-discrim} presented a limited usage of automated design with collective perception, due to a highly specific task and setup, with very little freedom for artificial evolution (only parameters of pre-designed controller rules were optimized). Another example for automated design on a controller with collective perception is in \cite{Shaukat2016-sourcelocal}, where an underwater swarm could locate an radio signal source and then move towards it as a cohesive group. Unfortunately, these real-life applications --although showing successful collective perception-- are limited in their automated design as they only optimize pre-designed control parameters (a biased controller with model parameters). A much more expressive optimization was shown by \cite{floreano2007evolutionary}, where a swarm of robots, controlled by evolved neural networks (which are model agnostic approximators), learned to locate the food in their environment by emerging a communication scheme. This study showed that swarm behavior can develop with model-agnostic  controllers undergoing evolution, even if such a behavior is not directly specified in the fitness function. 

More similar to the work presented here, \cite{ramos2019evolving} and \cite{witkowski2016emergence} presented automated design approaches to optimize not only controller parameters but controllers themselves (neural networks in these particular examples). \cite{ramos2019evolving} showed that a simulated swarm could learn similar flocking behavior, when directly optimizing for fitness functions that closely relate to well-known flocking rules (e.g. \cite{reynolds1987flocks}). In our opinion, these results are unsurprising as the flocking is directly optimized for, rather than emerged. Additionally, such an approach can be unsuitable for robots, as it may limit the controller's robustness when conditions change and with possible failures \cite{floreano2000evolutionary}. In our opinion, a more robust approach would be to favor a resultant swarm behavior of swarm (behavioral fitness) rather than directly rewarding the desired actions of individuals (functional fitness) \cite{floreano2000evolutionary}. \cite{witkowski2016emergence} showed that flocking-like behavior emerged when evolving a swarm of simulated agents only for exploiting a vital source for survival and reproduction. These results show that with an environment-driven task, swarm-level behavior can emerge if it provides an advantage with respect to solving the task collectively. In this paper, we hope to see the similar collective behavior emerge, by using a very generic controller (i.e. model agnostic) and optimizer in a more strict setting. 

To summarize, this study positions itself uniquely in that it hopes to show that, with limited sensing and true collective perception (no explicit communication or direct sensing), emergent collective behavior can be induced by the environment using a model-agnostic controller that undergoes evolution. 


\section{Methodology}
Full code base can be found in the following git repository: \\ \href{https://github.com/fudavd/EC_swarm/tree/GECCO_2022}{https://github.com/fudavd/EC\_swarm/tree/GECCO\_2022}

\subsection{Controller}
As a single behavior is considered during the evolutionary process, we use the same controller on every robot in the swarm. Thus, an individual in our EA refers to a single homogeneous swarm with the controllers of its constituents being an exact copy of the brain encoded by the genome. To be clear, the controller for each member of the swarm will be the same, but their respective input will likely differ due to their different positions and orientation within the swarm. Each controller provides a target velocity to the two wheels of the corresponding differential drive robot. For this we use a fully connected reservoir Neural Network (NN) controller where we only optimize the output layer using DE. The choice of this controller is based on two advantages: 1) NN are known to be model agnostic function approximators which means that there is no prior bias on the possible behaviours. 2) reservoir NN reduce the search space significantly (only the output layer is optimized) thereby increasing the learning rate.

In more detail, as an input to the reservoir NN we provide relative positional and heading information from 4 directional sensors, plus a local value sensor reading of a scalar field map (thus in total 9 different inputs) every 0.1s. The 4 directional sensors cover a combined 360 degrees view of the robot's surrounding, with a single sensor only sensing the nearest neighbour within a specific locally defined quadrant: Front-, Back-, Left-, and Right- quadrant (90 degrees each). The information obtained are the distance and relative heading of this neighbour with respect to the controlled robot (2 dimensions of information per sensor, in 4 quadrants). If no neighbour is present within the quadrant's maximum range (2m) the sensor set its distance measured to 2.0 and relative heading to 0. Lastly, local values of the scalar field are provided based on the robot's position in the environment. All sensor inputs are pre-processed to a [-1, 1] range, before being fed into the NN. The two NN outputs specify the heading- ($w \in [-1,1]$) and forward velocity ($v \in [0, -1]$), which are rescaled and translated to correct wheel speeds (up to $\pm \unit[14]{cm/s}$ per wheel). 

The reservoir NN architecture consist of an input layer ($\mathbf{s}_{in}$) with 9 neurons, 2 hidden layers (with Softplus functions) of the same size as the input layer, and an output layer (tanh$^{-1}$ of size 2). All weights are randomly initialized between [-2, 2] with a uniform distribution. We set all biases to 0 and will only optimize the weights of the output layer during evolution (18 weights). The final NN controller can be formalized as follows:

\begin{equation}
    NN = \mathrm{tanh}^{-1}\left(\mathbf{W}_{out}\mathrm{ReLU}\left(\mathbf{W}_{h2}\mathrm{ReLU}\left(\mathbf{W}_{h1}\mathbf{s}_{in}\right)\right)\right) 
    \tag*{$\textrm{with,} \quad  \mathbf{W}_{h1, h2} \in \mathbb{R}^{9\times9} \quad \textrm{and} \quad \mathbf{W}_{out} \in \mathbb{R}^{2\times9}$\newline}
\end{equation}

Each individual in the EA population receives the same reservoir ($\mathbf{W}_{h1, h2}$) in their NN controller but with varying output layers $\mathbf{W}_{out}$. We encode the reservoir NN controller's phenotype by a vector of weights with $2\times9=18$ floating point numbers (min/max value $\pm10.0$) as its genotype ($x$), specifying the rows in $\mathbf{W}_{out}$.

\begin{equation}
    x = \left[\mathbf{W}_{out_{1:}}, \mathbf{W}_{out_{2:}}\right] 
    \tag*{$\textrm{with,} \quad x \in \mathbb{R}^{18}$}
\end{equation}

\begin{figure}[hpt!]
  \includegraphics[width=0.9\linewidth]{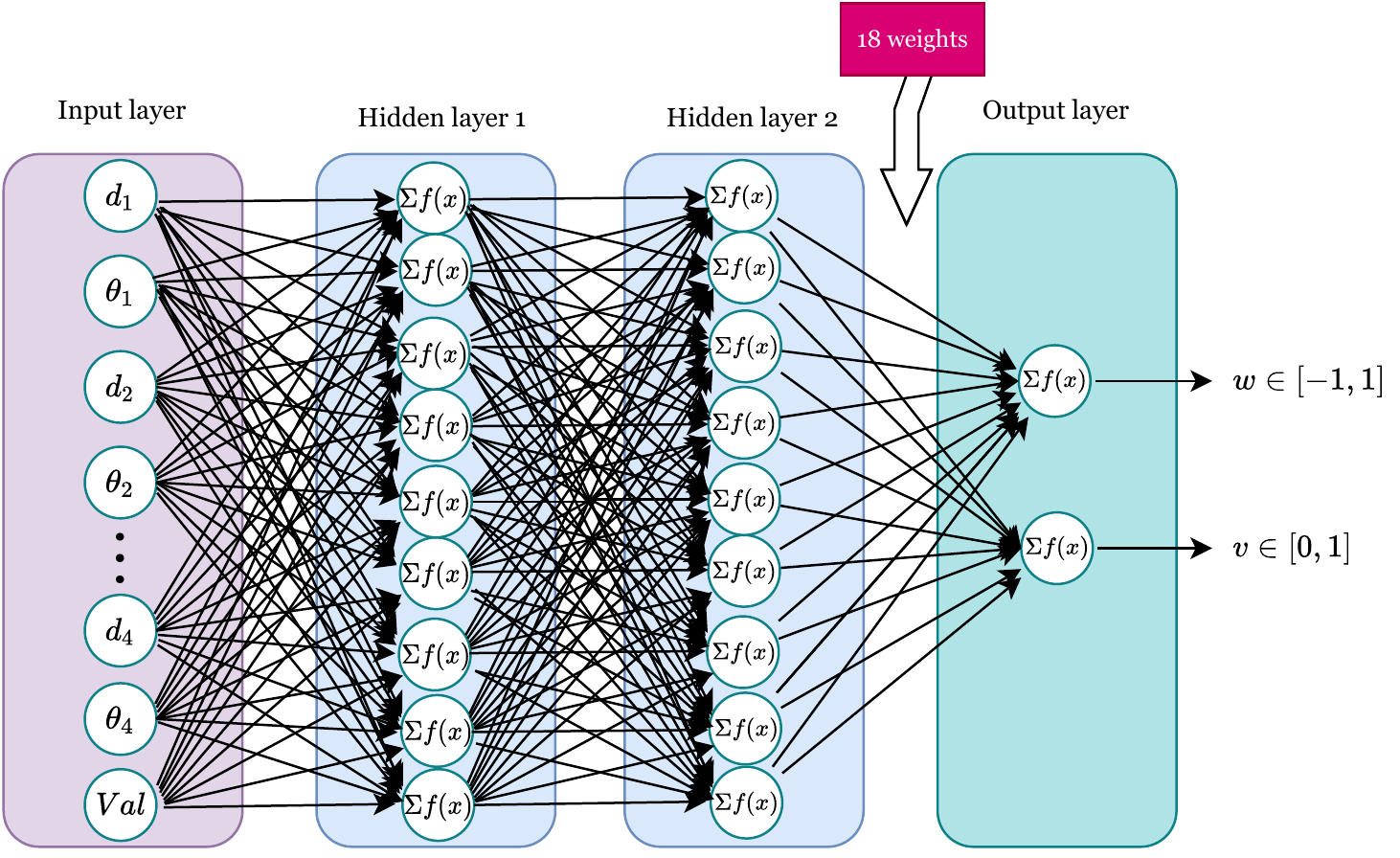}
  \centering
  \caption{Phenotype of the controller - Neuron Network}
  \label{fig:NN}
\end{figure}

The phenotype of the controller is illustrated in \autoref{fig:NN}.

\subsection{Evolutionary Algorithm}
Evolutionary methods for solving NP-hard optimization problems have become a very popular research topic in recent years. Some of the more established methods are genetic algorithm (GA), particle swarm optimization (PSO), and differential evolution (DE), which all have shown to be able to solve hard optimization problems for various domains \citep{Deng2006,Ai2009,Wisittipanich2011}. \citep{Kachitvichyanukul2012} states that GA is more well-established because of its much earlier introduction however it has less ablility to reach good solution without local search, while the more recent PSO and DE algorithms have started to attract more attention especially for continuous optimization problems. However in PSO, the best particle in the swarm exerts its one-way influence over all the remaining solutions in the population. This often leads to premature clustering around the best particle, especially if the fitness gaps are large. This is definitely an unwanted property, therefore in this paper, we choose DE to search our best solution. DE is a population-based Evolutionary Algorithm (EA) that samples new candidates by perturbing the current population \cite{Storn1997}. The three main components in this method are as follows:

Differential mutation operator: a new candidate is generated by randomly picking a triplet from the population, $(x_i,x_j,x_k)\in X$, then $x_i$ is perturbed by adding a scaled difference between $x_j$ and $x_k$, that is:
        \begin{align}\label{eq:de1}
              y = x_i + F(x_j-x_k)
        \end{align}
where $F\in R_+$ is the scaling factor. 

Uniform crossover operator: sample a binary mask $m \in \{0, 1\}^D$ according to the Bernoulli distribution with probability p = P(md = 1) shared across all D dimensions, and calculate the final candidate according to the following formula:
        \begin{equation}\label{eq:de2}
              v = m \odot y+(1-m) \odot x_i
        \end{equation}

The last component is a selection mechanism: the authors of \cite{Storn1997} proposed to use the “survival of the fittest” approach, i.e., combine the previous population with the new one and select N candidates with the highest fitness values, i.e., the deterministic ($\mu$ + $\lambda$) selection.

Studies \cite{Panda2008,Tomczak2020,Luo2021} have shown that the schemes for the trial vector (candidate) generation can have a significant influence on the algorithm's performance. Here we follow general recommendations which can be found in literature \cite{Pedersen2010} for stable exploration behavior, namely the crossover probability (CR) being fixed to a value of 0.9 and the scaling mutation factor (F) being fixed to a value of 0.5 (see \autoref{tab:parameters}). 

\begin{figure*}[hpt!] 
  \centering
  \includegraphics[width=400pt]{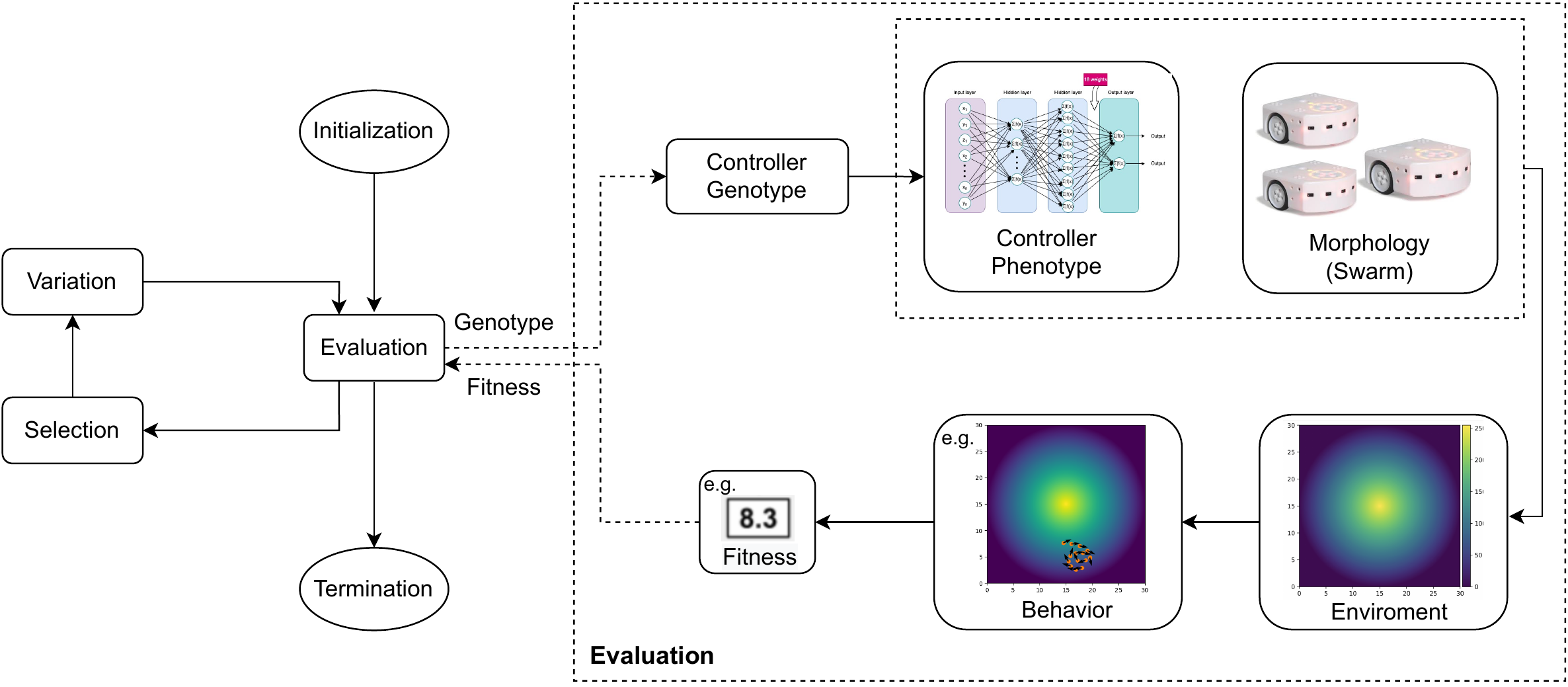}
  \vspace{-1em}
  \caption{Differential Evolution Framework: This is a DE framework to change the weights of the NNs of the robots to improve their controllers for the tasks. In the Evaluation box, we show examples of a robot morphology, controller, environment, behavior and fitness value.}
  \label{fig:whole_process}\vspace{-1em}
\end{figure*}

We apply DE to change the weights of the NNs of the robots to improve their controllers for the tasks. The whole process is illustrated in \autoref{fig:whole_process}. The pseudocode of DE for evolving the controllers is shown in Algorithm \ref{alg:DE}.

\begin{algorithm}[ht!]
\footnotesize
  \caption{Differential Evolution}
  \label{alg:DE}
  \begin{algorithmic}[1]
    \State INITIALIZE population (X controllers w/ vectors of weights)
    \While{iter < MAXiter}
        \For{i=1;i<=X; i++}
            \State GENERATE three individuals $x_i,x_j,x_k$ from X randomly
            \State MUTATION using formula: $y = x_i + F(x_j-x_k)$
            \State CROSSOVER using formula: $v = m \odot y+(1-m) \odot x_i$
            \State EVALUATE controllers (based on fitness value) 
            \State SELECT survivors / UPDATE population
        \EndFor
    \EndWhile
 \end{algorithmic}
\end{algorithm}

\subsection{Performance measures}
\subsubsection{Fitness function}
\begin{figure}[h!]
\vspace{-0.5em}
  \centering
\begin{minipage}[c]{0.95\linewidth}
\centering
        \begin{minipage}[t]{.4\linewidth}
            \centering
            \subfloat[(a) Scalar field map]{\includegraphics[width=\textwidth]{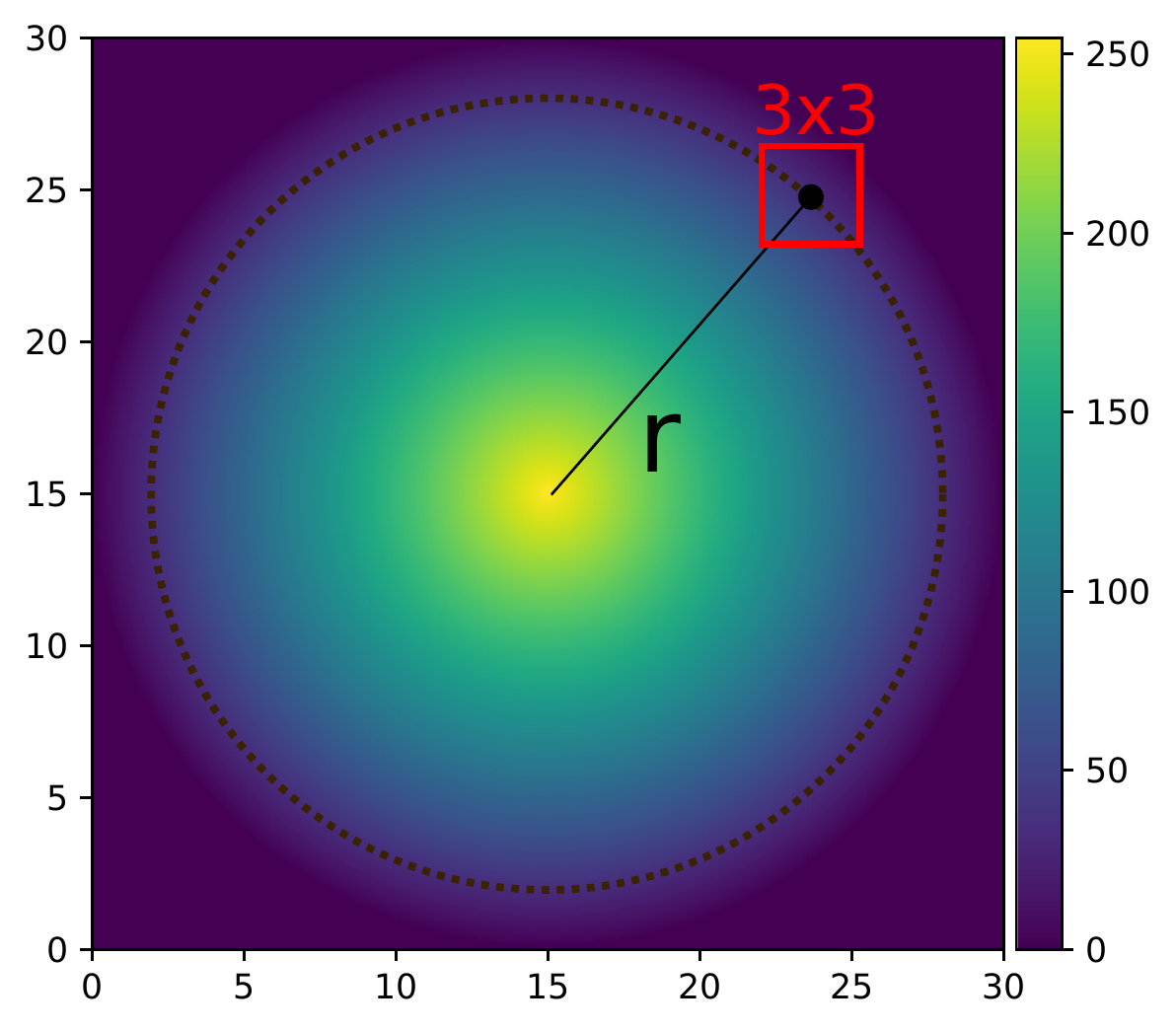}}
        \end{minipage}
\hspace{2em}
        \begin{minipage}[t]{.35\linewidth}
            \centering
            \subfloat[(b) Isaac gym]{\includegraphics[width=\textwidth]{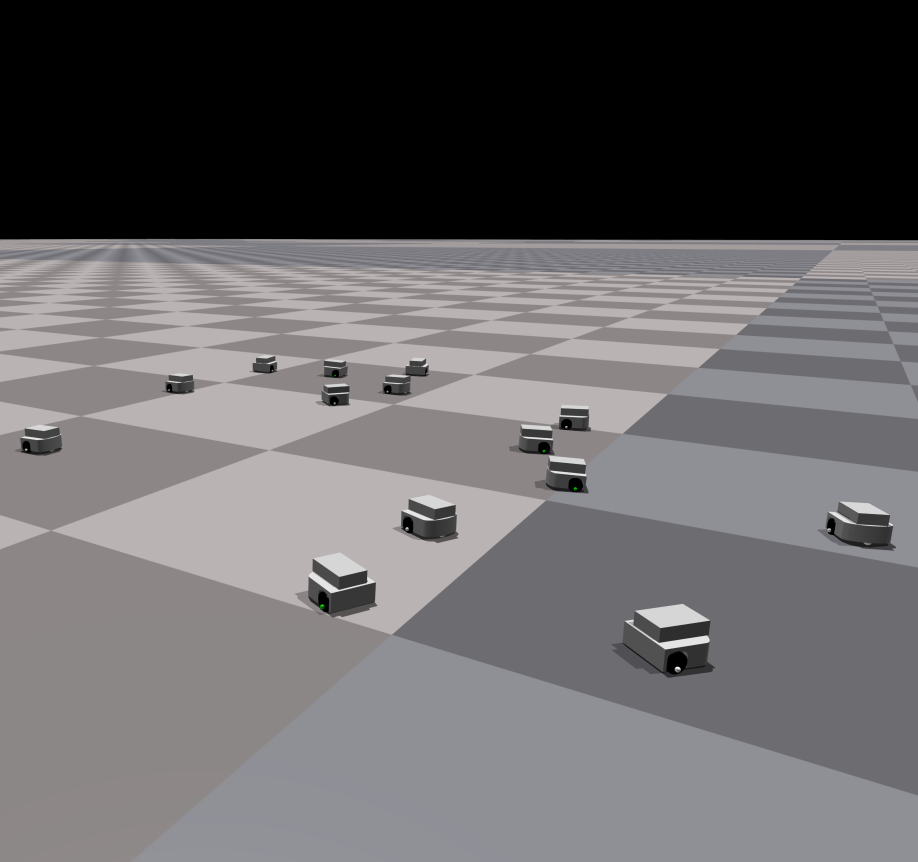}}
    \end{minipage}
\end{minipage}\vspace{-1em}
\caption{The swarm environment. (a) the scalar field map with its maximum value (255) in the center. Initial swarm position (shown as a black dot) is randomly placed on a circle with distance r away from the center. Individuals are randomly placed within a 3x3m bounding box (shown in red). (b) a random initial state is shown in simulation.}
\label{fig:swarm_env}
\vspace{-1em}
\end{figure}

In a homogeneous swarm, we evaluate fitness on group level. In this research, we are mainly interested in a pure evolution-driven effects on swarm behavior without directly encouraging any form of communication, collaboration or collectiveness in the evolutionary framework. Therefore, the fitness was solely dependent on the swarm's ability to follow the increasing gradient of the scalar field which is defined in the environment (shown in \autoref{fig:swarm_env}-a). Additionally, we would like to emphasize that our fitness function does not distinguish between robots that are sensing and following the increasing gradient collectively or as solitaries. The fitness function $F$ is defined as follows:

\begin{equation}
{F} = \frac{\sum_{t=0}^{T}{f_t}}{G_{max}\cdot{T}}\quad\text{and}\quad{f}_t = \frac {\sum_{n=1}^{N}{G_n}}{N}
\label{eq:r_i_sum}
\end{equation}

Where $G_n$ is the scalar value of the grid cell in which agent $n$ (of all agents, $N$) is located at a time $t$. Thus the fitness at a specific time ($f_t$) is calculated as the mean scalar value of all swarm members. Final fitness ($F$) is calculated by averaging all $f_t$ over total simulation time $T$. At last, we normalize using the maximum scalar value $G_{max}$, always equal to $255$ for all experiments. 

\subsubsection{Behavioral measures}
Besides our fitness function $F$, we want to investigate the possible emergence of flocking behavior. For this, we analyse two additional behavioral measures. The first measure: \textit{order} ($\Phi$), is defined as follows: 

\begin{equation}\label{eq:align}
{\Phi} = \frac{\sum_{n=1}^{N}{\varphi_n}}{N}\quad\text{and}\quad{\varphi_n}=\frac{\Big|\Big|\left(\sum_{p=1}^{P}{\angle{e^{j \theta_p}}}\right)+\angle{e^{j \theta_n}} \Big|\Big|}{P+1}
\end{equation}

Here, $\varphi_n$ defines the order value calculated for agent $n$. Which is the average current heading direction of agent $n$ (noted as $\angle{e^{j \theta_n}}$) and all its perceived neighbors $P$ (noted as $\angle{e^{j \theta_p}}$). The total swarm order $\Phi$ is then defined as the average $\varphi_n$ over all agents in the swarm. Order measure gives a powerful insight about the alignment of agent's direction of motion. If all agents move towards the same direction, then the order measure approaches to 1; and if they move in different directions, order approaches to 0.

In addition to order, the number of collisions between robots are also checked periodically. This check is not trivial since robots are modelled in a physics simulator, all collisions are realistically modeled and certainly have affect on the movement capabilities of colliding robots. In other words, when agents are collided, they can tip over, stop or drag each other because of the physical interaction between them. We will report on collision by plotting the total number of collision present in the swarm as a function of time.

The final measure to analyze is the trajectories of robots during an experiment. By including and highlighting the start and end points as well, presenting the trajectories of robots on the scalar field provides a clear intuition about the swarm's motion characteristics. Trajectories can show us if there is any correlated forward motion or rotation among the swarm and the relative difference of the local values of the scalar field between start and end points. In addition, trajectories can reveal any kind of behavioral differences between controllers that are evolved in different conditions and obtaining similar fitness. Finally visual inspection of the trajectories will be made. 

\subsection{Experimental setup}
\subsubsection{Simulator}
At the start of an evaluation, a single swarm is randomly placed at a fixed distance away from the center of the arena (sampled uniformly on a circle with radius r). Each swarm member is then randomly placed inside a 3x3m box (the red square in \autoref{fig:swarm_env}-a) around this position with a random heading. We simulate the behavior swarm for 10 minutes ($\mathrm{dt=0.05s}$) and sample the controller every 0.1s for new motor inputs. Simulation of the swarm is done using Isaac Gym (see \autoref{fig:swarm_env}-b), which allows us to simulate the whole population of 25 different swarm individuals in parallel. To reduce the effect of lucky runs, we evaluate every individual 2 times, in which the lowest value is taken as the final fitness. Thus, in total we evolve 25 individuals with 2 repeated tests for 100 generation in 30 different runs with 3 different evolutionary experiments (45.000 evaluations of 10 minutes in simulation).

\subsubsection{Evolutionary Experiment}
In total we perform DE optimization 3 times in different environments (3 arenas), all with the same task of collective gradient sensing. The individuals in the population are not a single robot (which is more conventional in evolutionary robotics) but rather describe a homogeneous swarm. An initial population of 25 individuals (i.e. 25 different homogeneous swarms) is randomly generated as the first generation. We test each individual twice and use the result of the lowest fitness. In each generation, 3 individuals are randomly picked to generate a new individual by crossover and mutation. The next generation is formed by the top 25 individuals from the previous generation plus the new candidates according to fitness value. The evolutionary process is terminated after 100 generations. All the evolutionary experiments are repeated 30 times independently to get a robust assessment of the performance per data set.

   
   



\subsubsection{Collective behavior, Flexibility \& Scalability}
After the three evolutionary experiments we re-test the single best controller over all 30 runs per arena (thus in total three different best controller). The best controller is tested in the same environment to evaluate if collective motion has emerged. We look at three different aspects of collective behavior over time: \textit{alignment} in terms of order (\autoref{eq:align}, $\Phi$), \textit{performance} as mean scalar value of the swarm (\autoref{eq:r_i_sum}, $f_t$), and \textit{collision} counted at every controller update. These metrics are plotted with respect to time.

Additionally, we investigate two types of swarm properties, namely flexibility and scalability \cite{csahin2004swarm}. Firstly, we (cross-)validate the overall best controllers per environment, meaning that the same swarm was re-tested in its own environment and the other two arenas for comparison. A similar performance in the other environments indicates flexibility of the found solution since changing environment sizes indicate a slower or faster change in the local values of the gradient for the same amount of displacement. Since the best controllers for each environment size is evolved to be the best with the corresponding rate of change of local values, different environment size and different rate of change of local values can be regarded as environmental variations. Secondly, to see the scalability of our control solution, we re-tested the overall best controller per arena, in the same arena with different swarm sizes (here we resized the initialization box accordingly). For both flexibility and scalability experiments we test the best controller 30 times per condition. 
The experimental parameters we used in the experiments are described in \autoref{tab:parameters}.
\begingroup

\renewcommand{\arraystretch}{0.6} 

\begin{table}[htp!]
\footnotesize
\caption{Main experiment parameters}
\centering
\vspace{-1em}
\begin{tabular}[b]{{p{0.20\linewidth}| p{0.15\linewidth}| p{0.5\linewidth}}}
\toprule
\textbf{DE}         & Value & Description \\ \midrule
~Swarm size 			 & ~14     & ~Number of robots in a swarm 	\\
~Arena size 			 & ~10/30/45     & ~Swarm environment 	\\

~Population size  & ~25    & ~Number of individuals (swarms) per generation     \\
~Generations      & ~100   & ~Termination condition for each run             \\ 
~Mutation (F)         & ~0.5   & ~Mutation factor for individuals        \\ 
~Crossover (CR)         & ~0.9   & ~Probability of crossover for individuals        \\ 
~Evaluation time  & ~10    & ~Duration of an evaluation in minutes \\ 
~Repetitions      &  ~30    & ~Number of repetitions per experiment \\ 
\bottomrule 
\toprule
\textbf{Flexibility}         & Value & Description \\ \midrule
~Swarm size 			 & ~14     & ~Number of robots in a swarm 	\\
~Arena size 			 & ~10/30/45     & ~Swarm environment 	\\
~Evaluation time  & ~10    & ~Duration of an evaluation in minutes \\ 
~Repetitions      &  ~30    & ~Number of repetitions per experiment \\ 
\bottomrule 
\toprule
\textbf{Scalability}         & Value & Description \\ \midrule
~Swarm size 			 & ~5/14/50     & ~Number of robots in a swarm 	\\
~Arena size 			 & ~10/30/45    & ~Swarm environment 	\\
~Evaluation time  & ~10    & ~Duration of an evaluation in minutes \\ 
~Repetitions      &  ~30    & ~Number of repetitions per experiment \\ 
\bottomrule 
\end{tabular}
\vspace{-1.5em}
\label{tab:parameters}
\end{table}
\endgroup

    

\section{Experiment Results}
All reported results and analysis data/scripts can be found in the following database: \href{https://doi.org/10.34894/IREUW1}{https://doi.org/10.34894/IREUW1}

\subsubsection{Efficacy}
We measure efficacy by the mean and maximum fitness averaged over the 30 independent runs at every generation. 

\begin{figure}[t!]
\vspace{-0.5em}
  \includegraphics[width=0.85\linewidth]{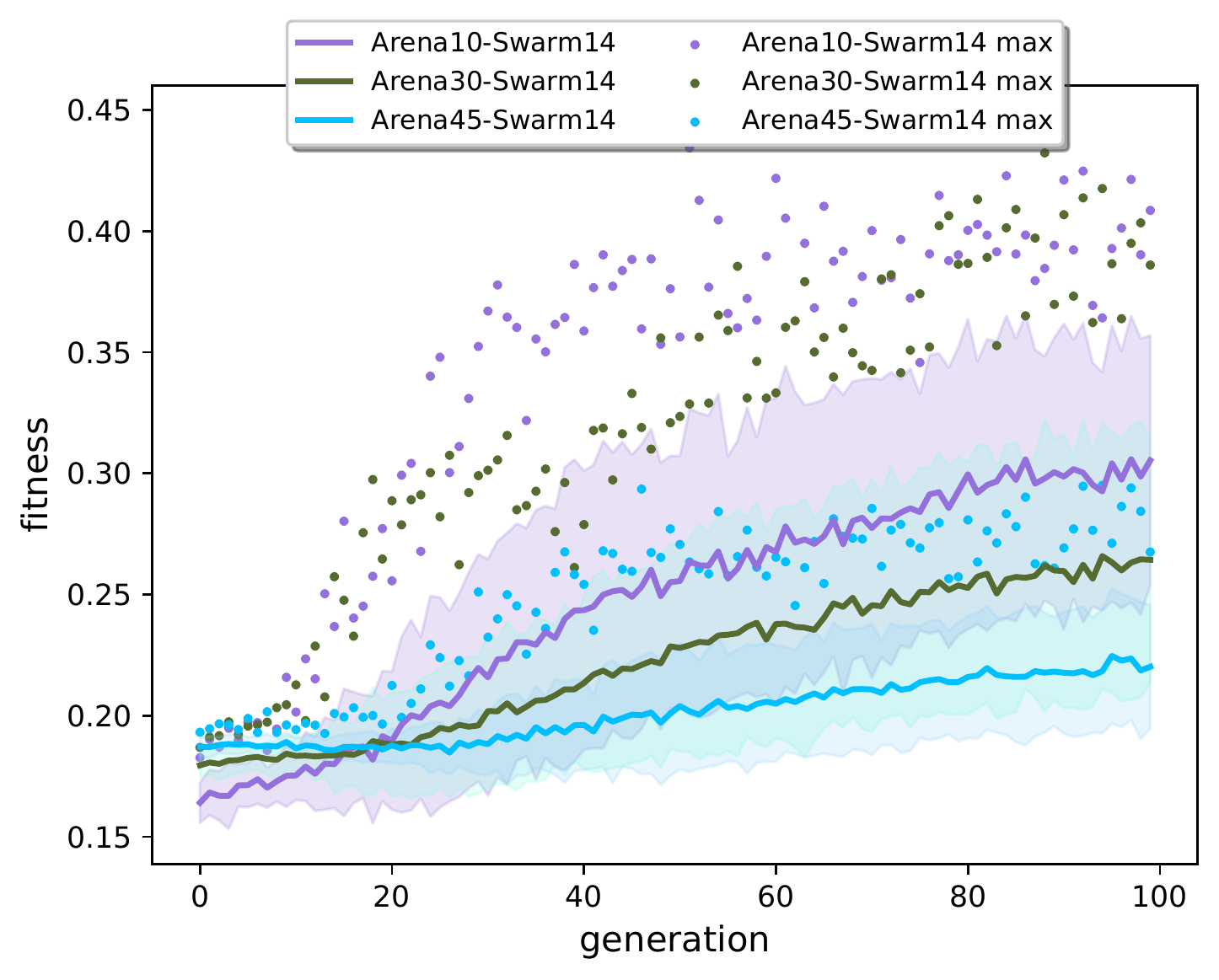}\vspace{-1.5em}
  \caption{The mean$\pm STD$ (line) and average max (dot) fitness over 100 generations (averaged over 30 runs) for all experimental condition (10x10: purple, 30x30: green, 45x45: blue.}
  \label{fig:fitness}
\end{figure}

For each evolutionary experiment with swarm size 14 we run the learning task 30 times at three different arenas. \autoref{fig:fitness} shows that the experiment in arena 45x45 has the highest average fitness at the start, followed by the experiment in arena 30x30. Then the fitness of all the evolutionary experiments increase steadily generation by generation. From around generation 28, the average fitness from arena 10x10 surpasses the other two, so does the max fitness. Similarly, around the same cross-over point, the average fitness from arena 45x45 becomes the lowest. In the end, the mean best solution over 30 runs were found in the 10x10 condition ($F_{10}=0.46 \pm 0.038$), followed by 30x30 ($F_{30}=0.39 \pm 0.055$), and 45x45 ($F_{45}=0.33 \pm 0.040$). Statistical analysis showed that these difference were signicant between all conditions ($p\ll 0.05$).

\begin{figure}[h]
\vspace{-0.5em}
  \includegraphics[width=0.65\linewidth]{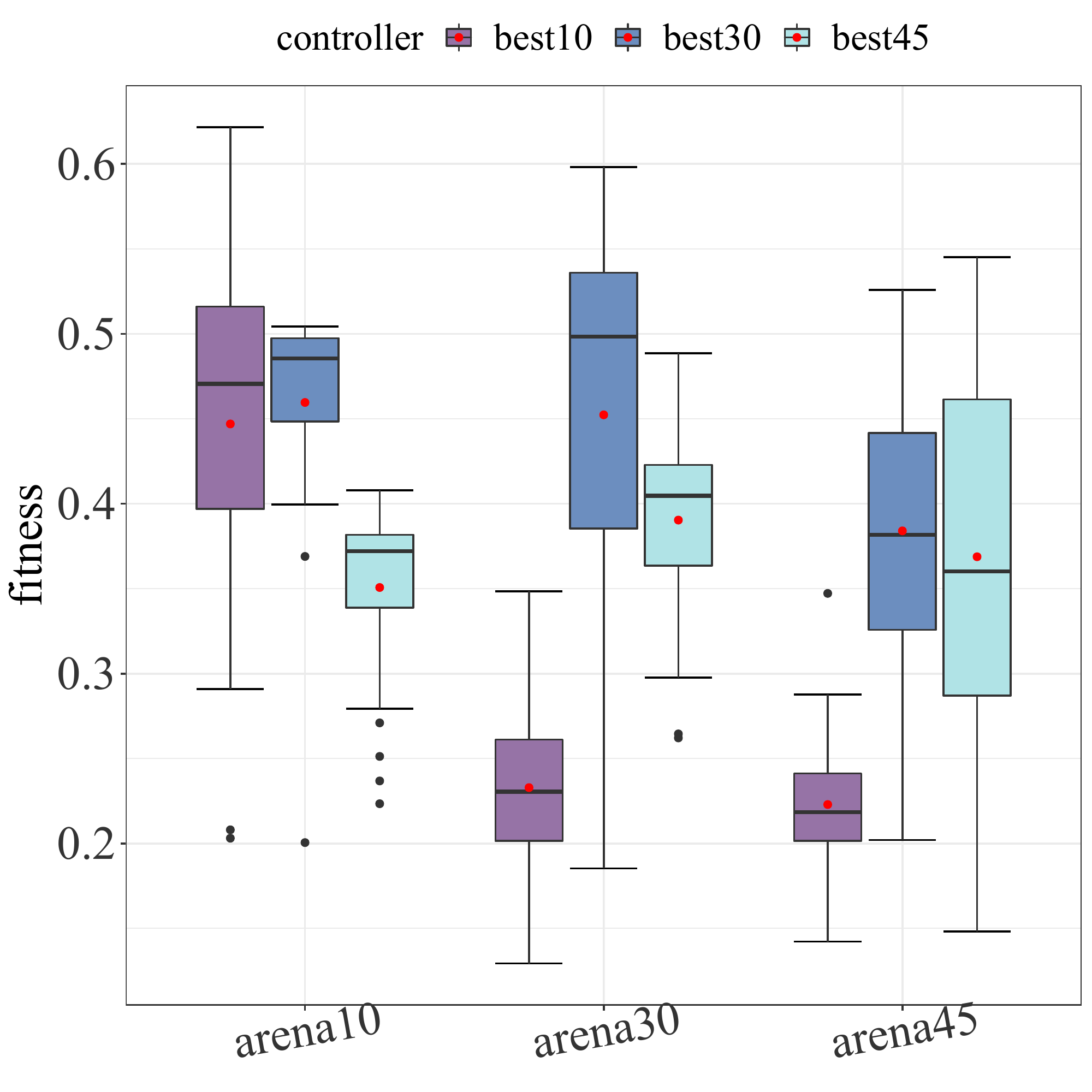}\vspace{-1em}
  \caption{Flexibility boxplot. Cross-validation of best controller in other arenas. Red dots show mean values.}
  \label{fig:cross_validation}\vspace{-0.5em}
\end{figure}

\begin{figure}[h!]
  \includegraphics[width=0.65\linewidth]{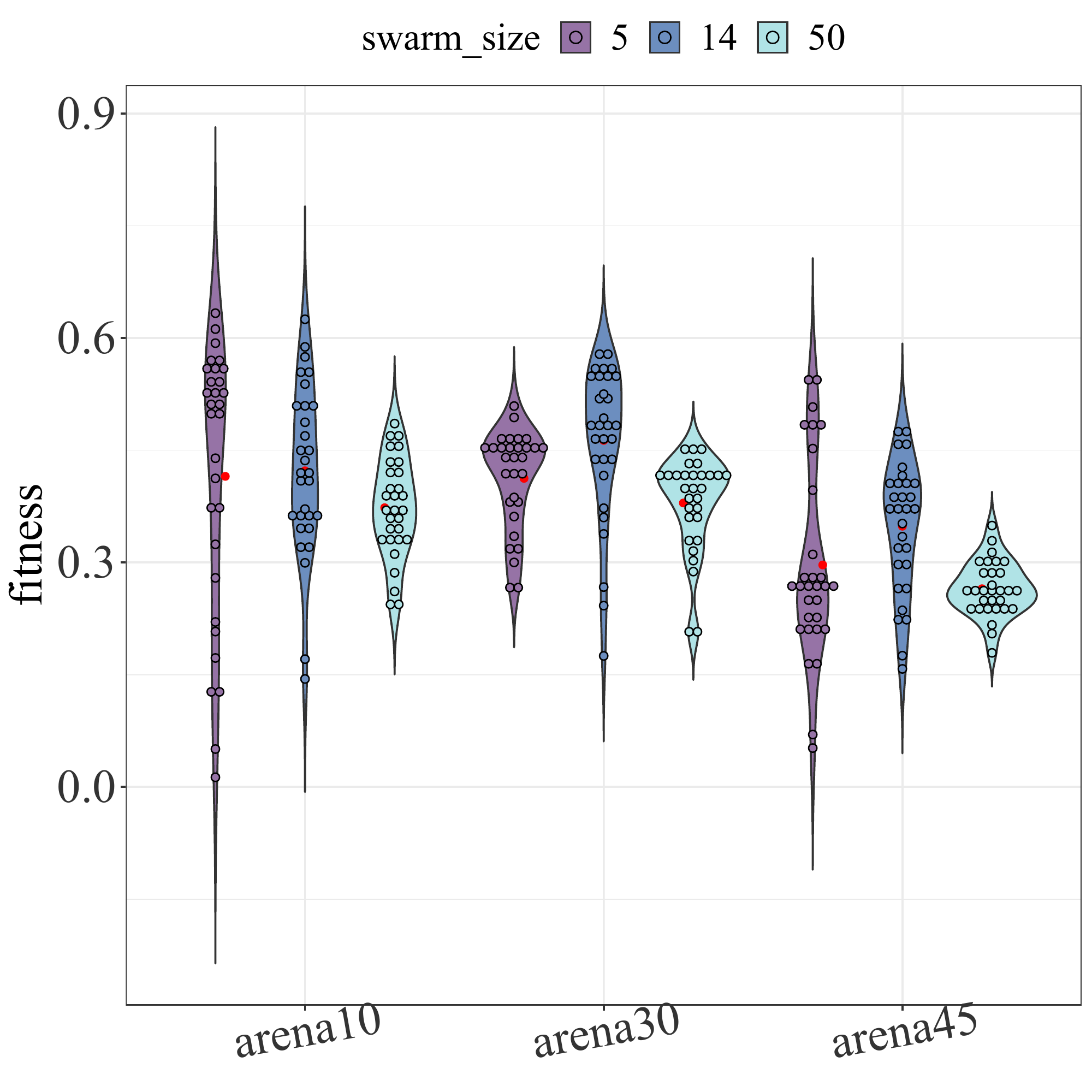}
  \vspace{-1.5em}
  \caption{Scalability violin plot. Re-test of best controllers with different swarm sizes (5: purple, 14: blue, 50: cyan) for 30 runs. Each circle represents a single run with red dots as mean value.
  }\vspace{-1em}
  \label{fig:scalability}
\end{figure}

\subsubsection{Flexibility}
The final swarm controller should be able to operate despite variations in its environment. We re-evaluate the overall best controllers' fitness from each evolutionary optimization (3 arenas with swarm size 14) in 9 additional tests across three different arenas, namely 10x10m, 30x30m and 45x45m, with 30 repetitions.

\autoref{fig:cross_validation} shows that the best controller from the 10x10 evolutionary run (in color purple) performs best in arena10, however it significantly drops in performance when re-tested in the other two larger arenas ($p < 0.05$). For the 30x30 run (blue boxplots), the best controller performs similarly well in both the smaller arena10 and its own (arena30), while the fitness is lower in the bigger arena45. The last best controller from the 45x45 run (cyan boxplots) has a relatively steady performance across all three arenas. From this plot, we may conclude that the best controller from the biggest arena 45x45 is the most flexible one. During evolution, the most flexible controller was evolved in the hardest environment. Apparently controllers evolved in larger arenas, perform relatively (equally) well in smaller arenas, while the reverse is not the case.

\subsubsection{Scalability}
The swarm should be able to operate with a wide range of group sizes and support large numbers of individuals without impacting performance considerably. That is, the coordination mechanisms and strategies to be developed for swarm robotic systems should ensure cooperation of the swarm with varying swarm sizes. To measure the scalability property, we choose the best controllers from evolutionary experiments (Arena10,30,45 with swarm size 14) and apply them on three different swarm sizes --being 5, 14, 50-- within the same environment.

\autoref{fig:scalability} shows that all the best controllers are quite scalable across three arena with close mean fitness. Experiment with swarm size 14 (blue violins) has the highest mean fitness across three arenas. Experiment with swarm size 50 has a relatively lower standard deviation which indicates a more robust performance. In other words, a bigger swarm size creates a bigger sample of the map (because it has a bigger coverage) which makes the estimation of the gradient better and the performance more stable. 

\subsubsection{Collective behavior}
\begin{figure*}
    \centering
    \begin{flushleft}{\textbf{A}} \end{flushleft}
    \vspace{-1em}
    \begin{minipage}[c]{0.9\textwidth}
        \begin{minipage}[t]{.225\textwidth}
            \centering
            \subfloat[10x10]{\includegraphics[width=\textwidth]{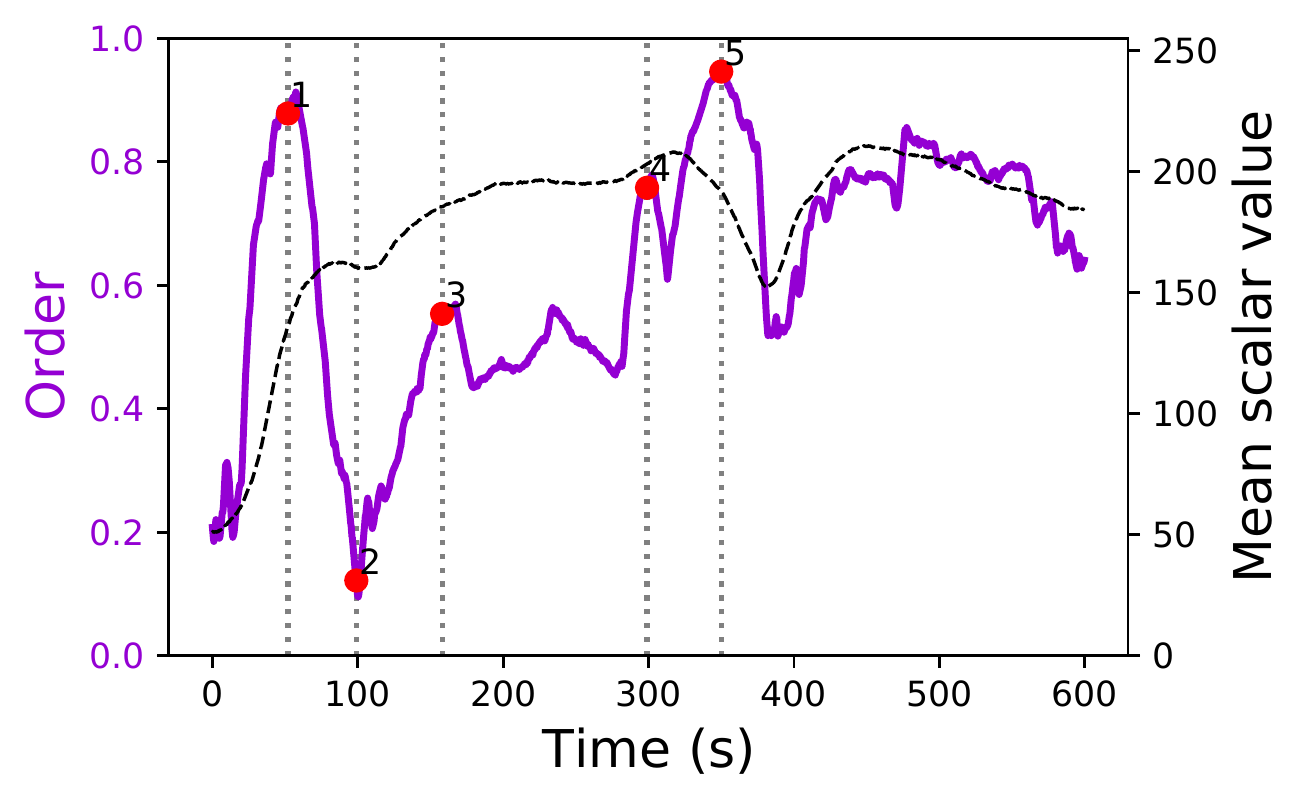}}
        \end{minipage}
        \hfill
        \begin{minipage}[t]{.15\textwidth}
            \centering
            \subfloat[1]{\includegraphics[width=\textwidth]{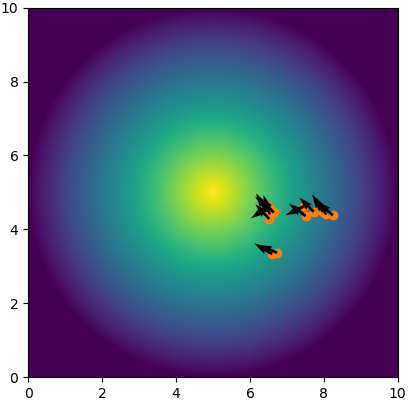}}
        \end{minipage}
        \hfill
        \begin{minipage}[t]{.15\textwidth}
            \centering
            \subfloat[2]{\includegraphics[width=\textwidth]{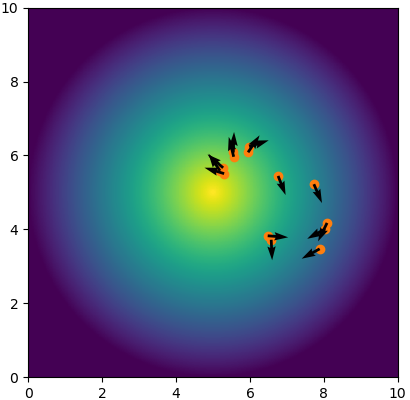}}
        \end{minipage}
        \hfill
        \begin{minipage}[t]{.15\textwidth}
            \centering
            \subfloat[3]{\includegraphics[width=\textwidth]{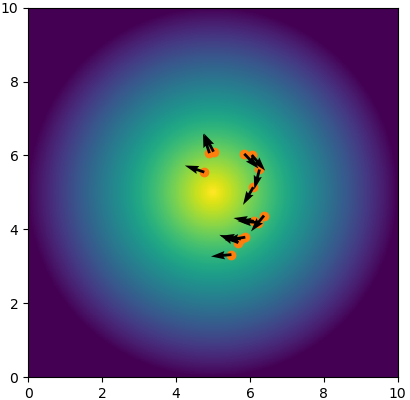}}
        \end{minipage}
        \hfill
        \begin{minipage}[t]{.15\textwidth}
            \centering
            \subfloat[4]{\includegraphics[width=\textwidth]{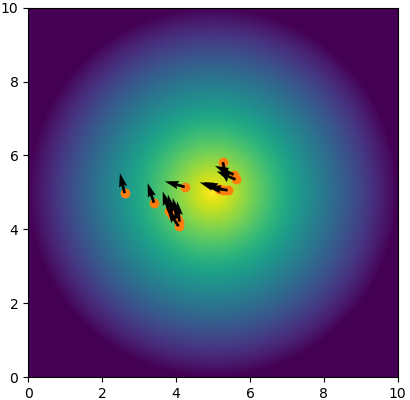}}
        \end{minipage}
        \hfill
        \begin{minipage}[t]{.15\textwidth}
            \centering
            \subfloat[5]{\includegraphics[width=\textwidth]{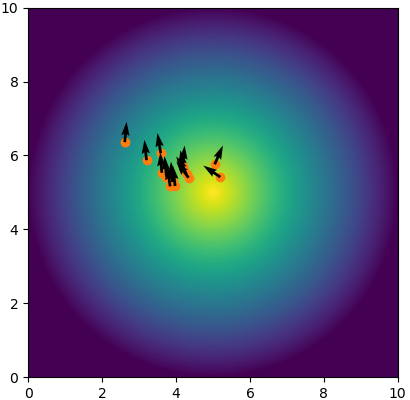}}
        \end{minipage}
\end{minipage}
\centering
    \begin{flushleft}{\textbf{B}} \end{flushleft}
        \vspace{-1em}
    \begin{minipage}[c]{0.9\textwidth} 
        \begin{minipage}[t]{.225\textwidth}
            \centering
            \subfloat[30x30]{\includegraphics[width=\textwidth]{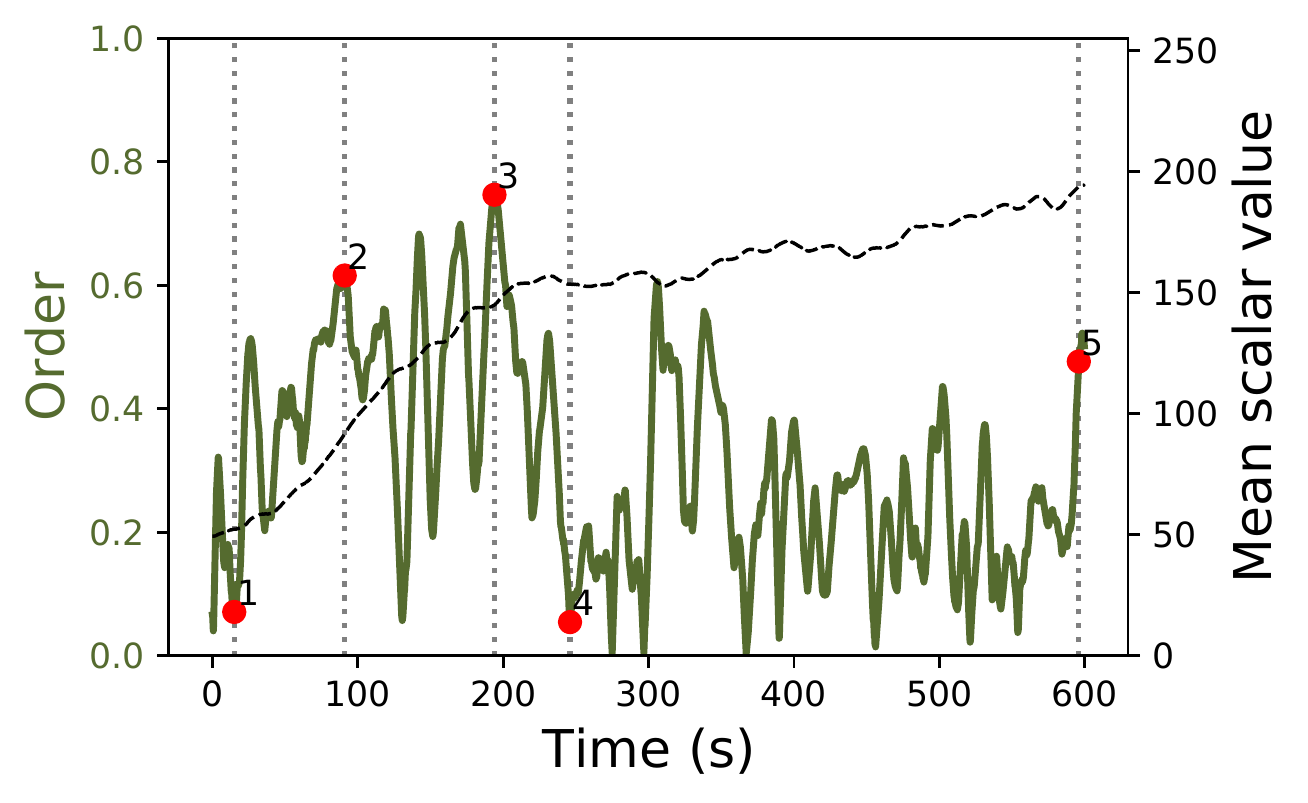}}
        \end{minipage}
        \hfill
        \begin{minipage}[t]{.15\textwidth}
            \centering
            \subfloat[1]{\includegraphics[width=\textwidth]{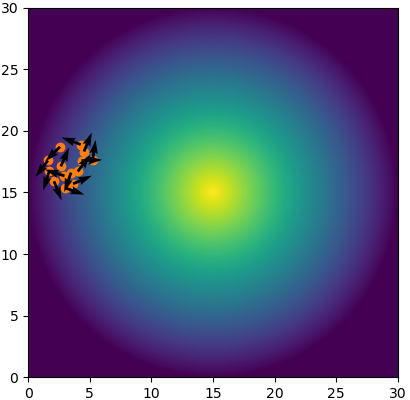}}
        \end{minipage}
        \hfill
        \begin{minipage}[t]{.15\textwidth}
            \centering
            \subfloat[2]{\includegraphics[width=\textwidth]{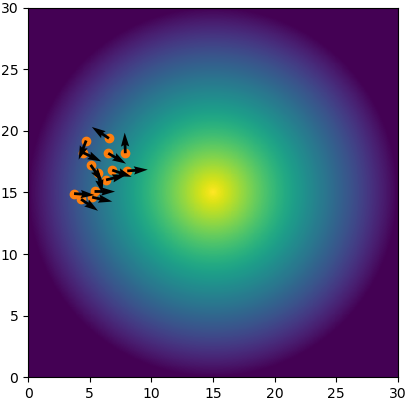}}
        \end{minipage}
        \hfill
        \begin{minipage}[t]{.15\textwidth}
            \centering
            \subfloat[3]{\includegraphics[width=\textwidth]{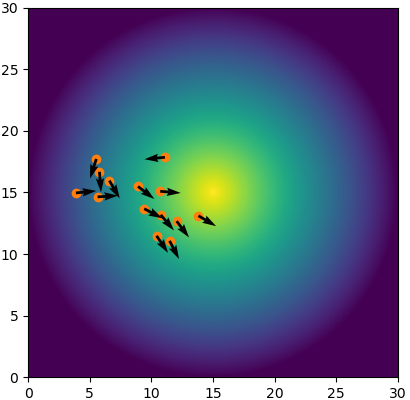}}
        \end{minipage}
        \hfill
        \begin{minipage}[t]{.15\textwidth}
            \centering
            \subfloat[4]{\includegraphics[width=\textwidth]{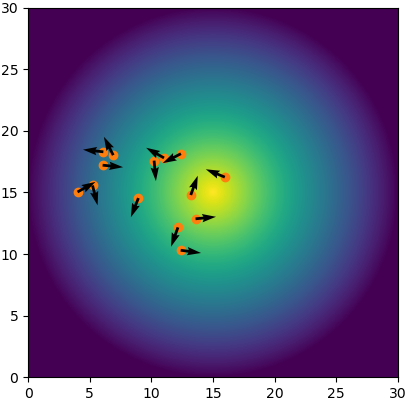}}
        \end{minipage}
        \hfill
        \begin{minipage}[t]{.15\textwidth}
            \centering
            \subfloat[5]{\includegraphics[width=\textwidth]{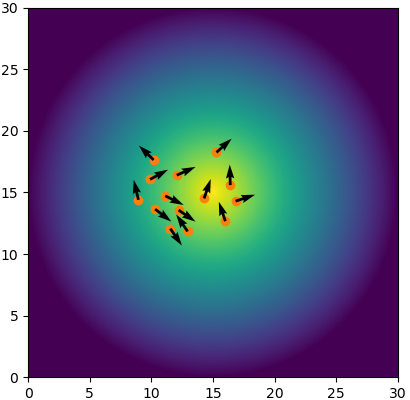}}
        \end{minipage}
\end{minipage}
\centering
    \begin{flushleft}{\textbf{C}} \end{flushleft}
    \vspace{-1em}
    \begin{minipage}[c]{0.9\textwidth} 
        \begin{minipage}[t]{.225\textwidth}
            \centering
            \subfloat[45x45]{\includegraphics[width=\textwidth]{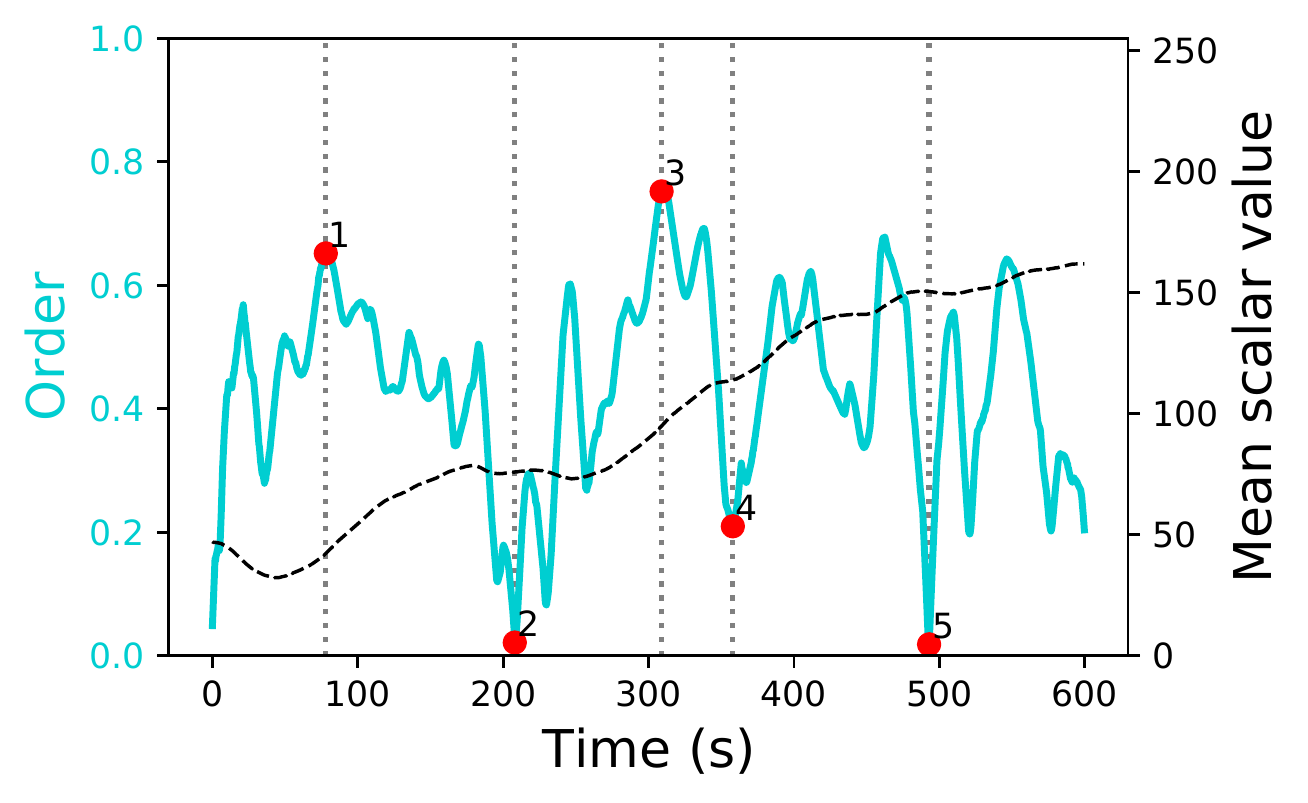}}
        \end{minipage}
        \hfill
        \begin{minipage}[t]{.15\textwidth}
            \centering
            \subfloat[1]{\includegraphics[width=\textwidth]{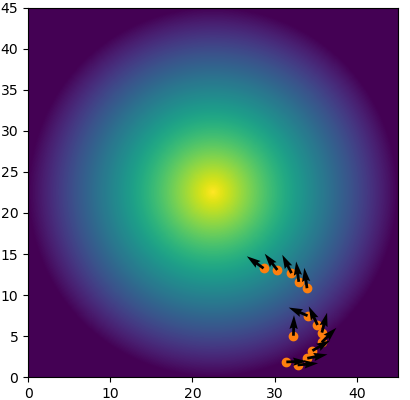}}
        \end{minipage}
        \hfill
        \begin{minipage}[t]{.15\textwidth}
            \centering
            \subfloat[2]{\includegraphics[width=\textwidth]{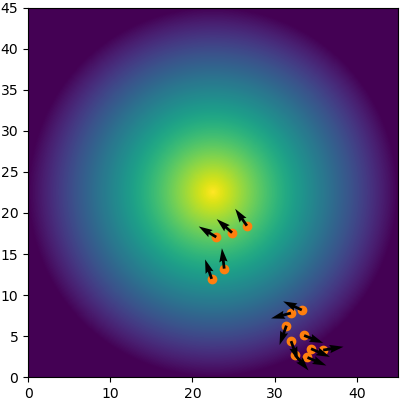}}
        \end{minipage}
        \hfill
        \begin{minipage}[t]{.15\textwidth}
            \centering
            \subfloat[3]{\includegraphics[width=\textwidth]{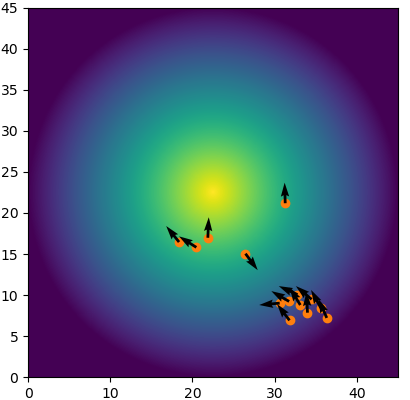}}
        \end{minipage}
        \hfill
        \begin{minipage}[t]{.15\textwidth}
            \centering
            \subfloat[4]{\includegraphics[width=\textwidth]{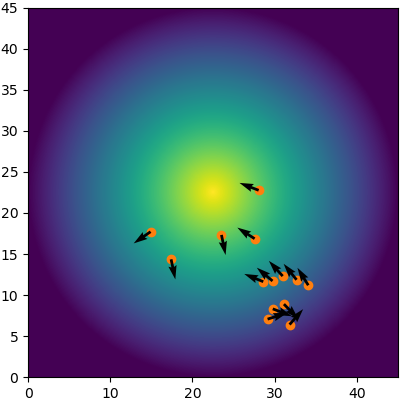}}
        \end{minipage}
        \hfill
        \begin{minipage}[t]{.15\textwidth}
            \centering
            \subfloat[5]{\includegraphics[width=\textwidth]{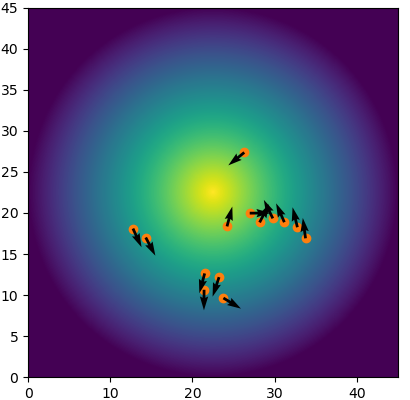}}
        \end{minipage}
\end{minipage}\vspace{-1em}
    \caption{Flocking behavior of the final controller re-tested in the same environment. The three line plots on the left show the correlation between order (alignment) and mean scalar value (performance). The snapshots on the right are corresponding to each line plot. The dashed black lines show the mean scalar value and the colored lines show the order value. Distinct peaks in alignment are highlighted with numbers 1-5. Please note the different scales on each map.}
    \label{fig:order}\vspace{-1em}
\end{figure*}

\autoref{fig:order} shows the correlation of alignment (Order, $\Phi$) within the swarm members and their performance (Mean scalar value, $f_t$) over time. We observe that when order peaks, i.e. namely red point 1, 3, 4 and 5 in arena 10x10m, red points 2, 3 and 5 in arena 30x30m and red points 1,3 in arena 45x45m, the mean scalar value increase (i.e. the tangent of the dashed black line is highly positive). Vice versa, when the colored lines reach to lower order value, the mean scalar value decrease or stay the same (i.e. tangent is near zero). 

The snapshots on the right are correspond with the same numbered points in the order/mean scalar value plot. We can see that in most cases when a swarm is becoming more aligned (i.e. order is high), the direction of movement is similar to the direction of increasing gradient (i.e. towards the center). In the low alignment case (i.e. low order) most of the members disperse in different directions, causing no net difference in the mean scalar value.

Impressively, here we can see a collective `search-like' behavior emerge that is reminiscent of an optimization algorithm trying to maximize a function. Scalar values are periodically and collectively sensed by the swarm. When the order is low, the swarm samples in different direction for higher values i.e. "exploration". In contrast, during the high peaks the swarm collectively moves in the direction of the gradient which causes the mean scalar value to increase i.e. "exploiting". These exploration exploitation states are periodically changed to escape local optima and find a global maximum. Even more exciting, when mean scalar value plateaus (i.e. swarm arriving near maximum) there is a clear switch in behavior and a new pattern emerges. In the 10x10 the swarm approaches the center and overshoots it, directly followed by a fast recovery toward the center. For 30x30 controller, the agents randomly spread around the maximum, slowly creeping inward. Unfortunately, the 45x45 experiment run was too short to see a final pattern emerge.

\begin{figure}[h]
     \centering
     \begin{subfigure}{0.2025\textwidth}
         \centering
         \includegraphics[width=\textwidth]{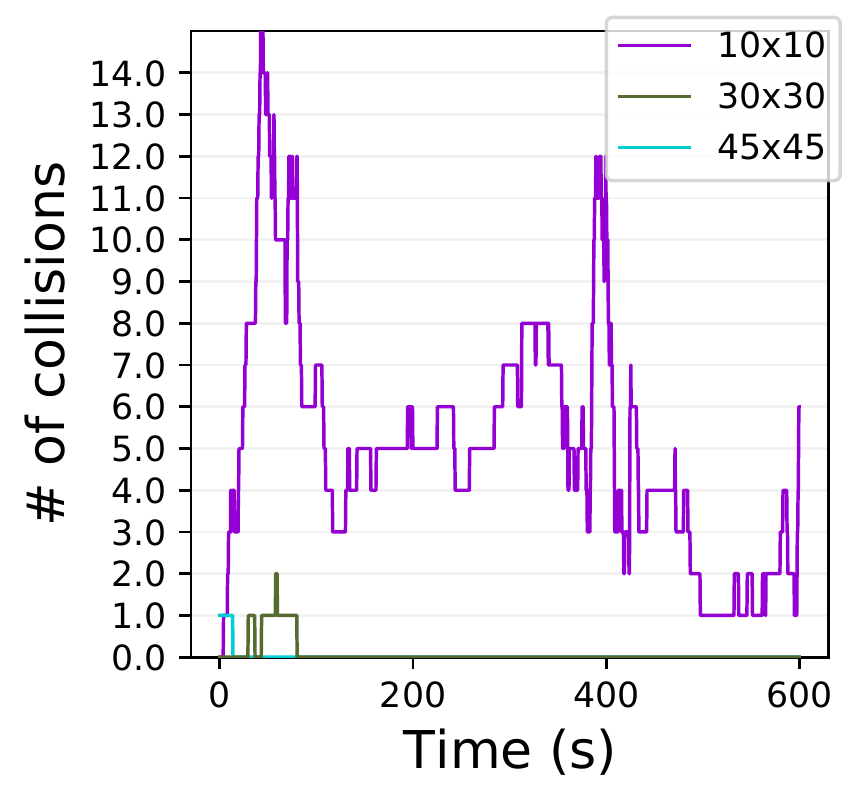}
         \caption{(a) Number of collisions}
         \label{fig:collis}
     \end{subfigure}
     \hfill
     \begin{subfigure}{0.20\textwidth}
         \centering
         \includegraphics[width=\textwidth]{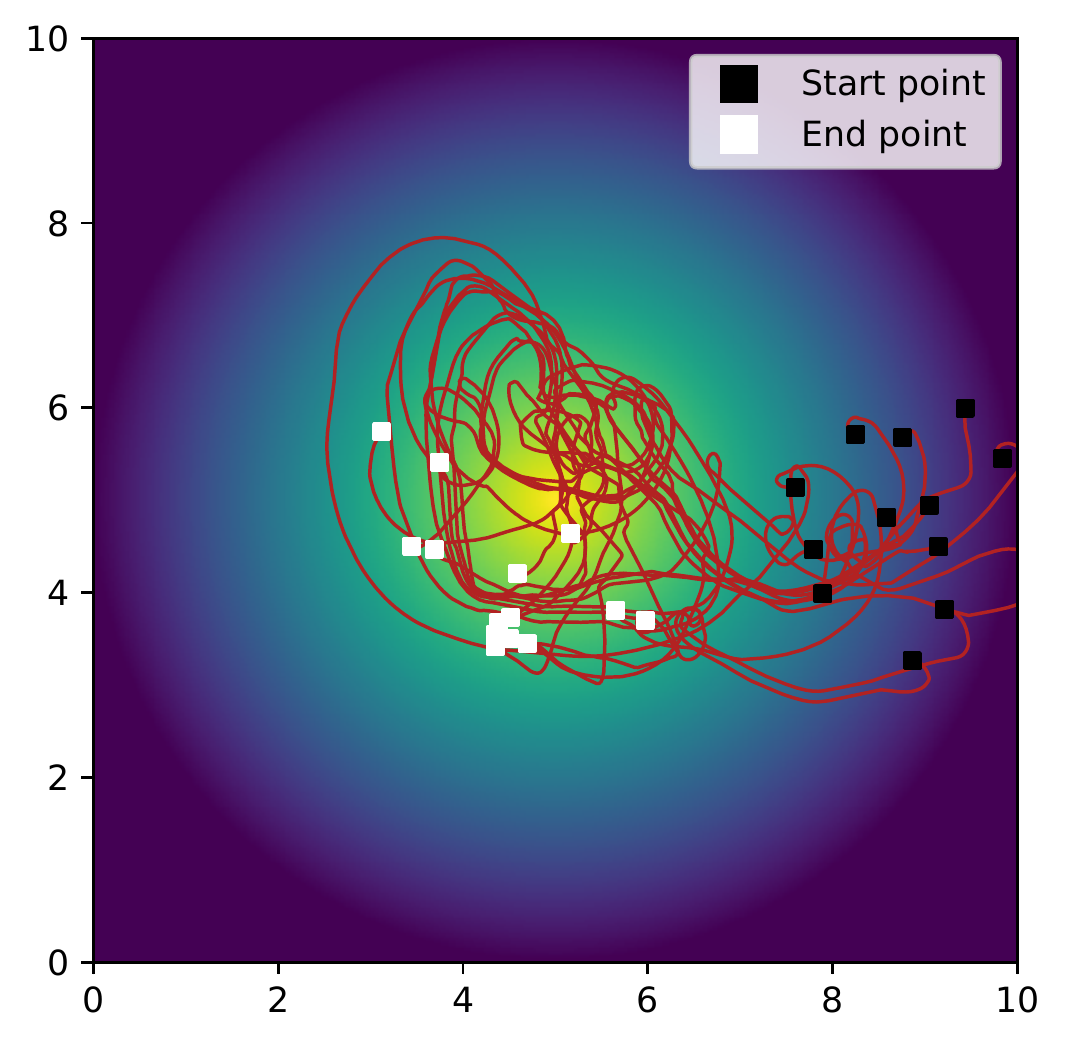}
         \caption{(b) 10x10}
         \label{fig:10x10}
     \end{subfigure}
     \begin{subfigure}{0.20\textwidth}
         \centering
         \includegraphics[width=\textwidth]{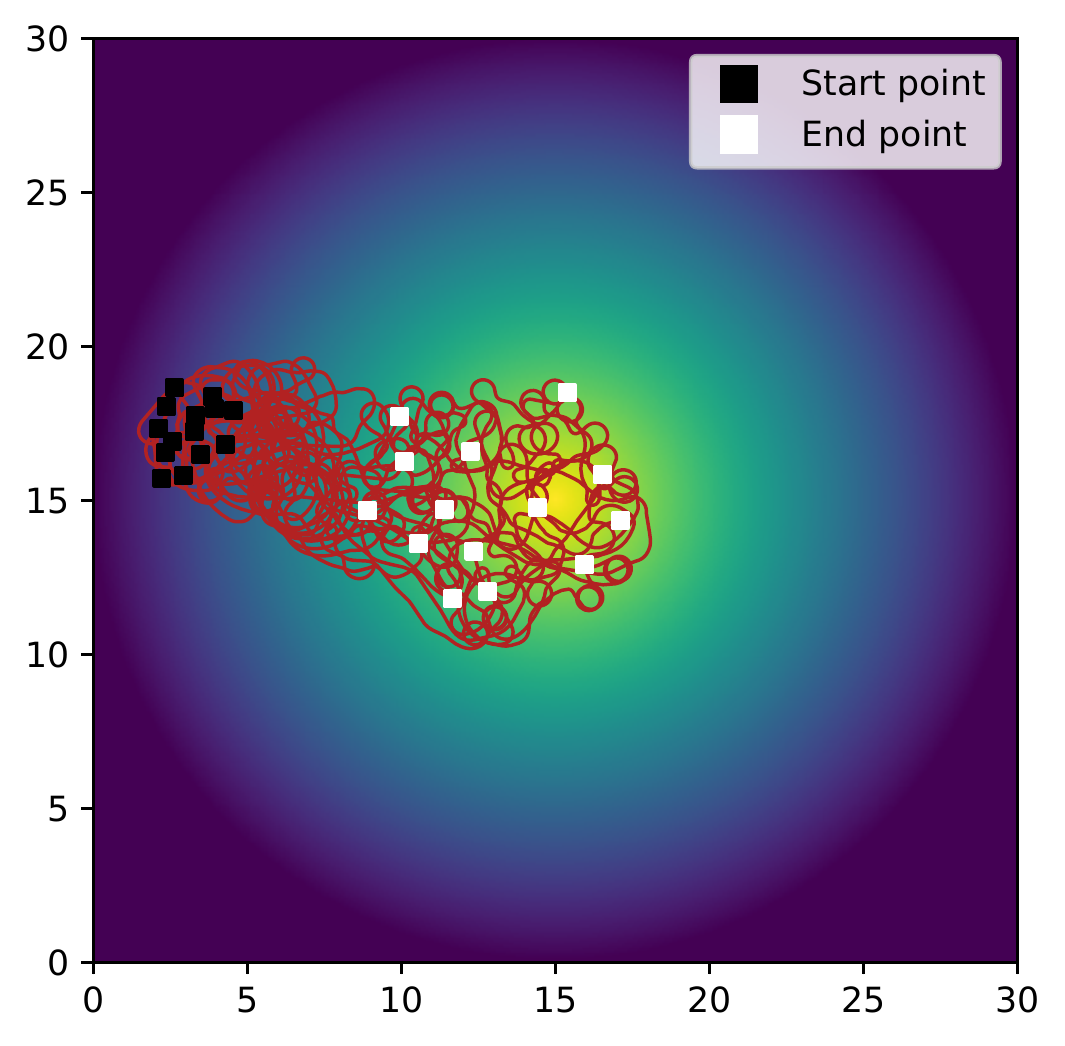}
         \caption{(c) 30x30}
         \label{fig:30x30}
     \end{subfigure}
     \hfill
     \begin{subfigure}{0.20\textwidth}
         \centering
         \includegraphics[width=\textwidth]{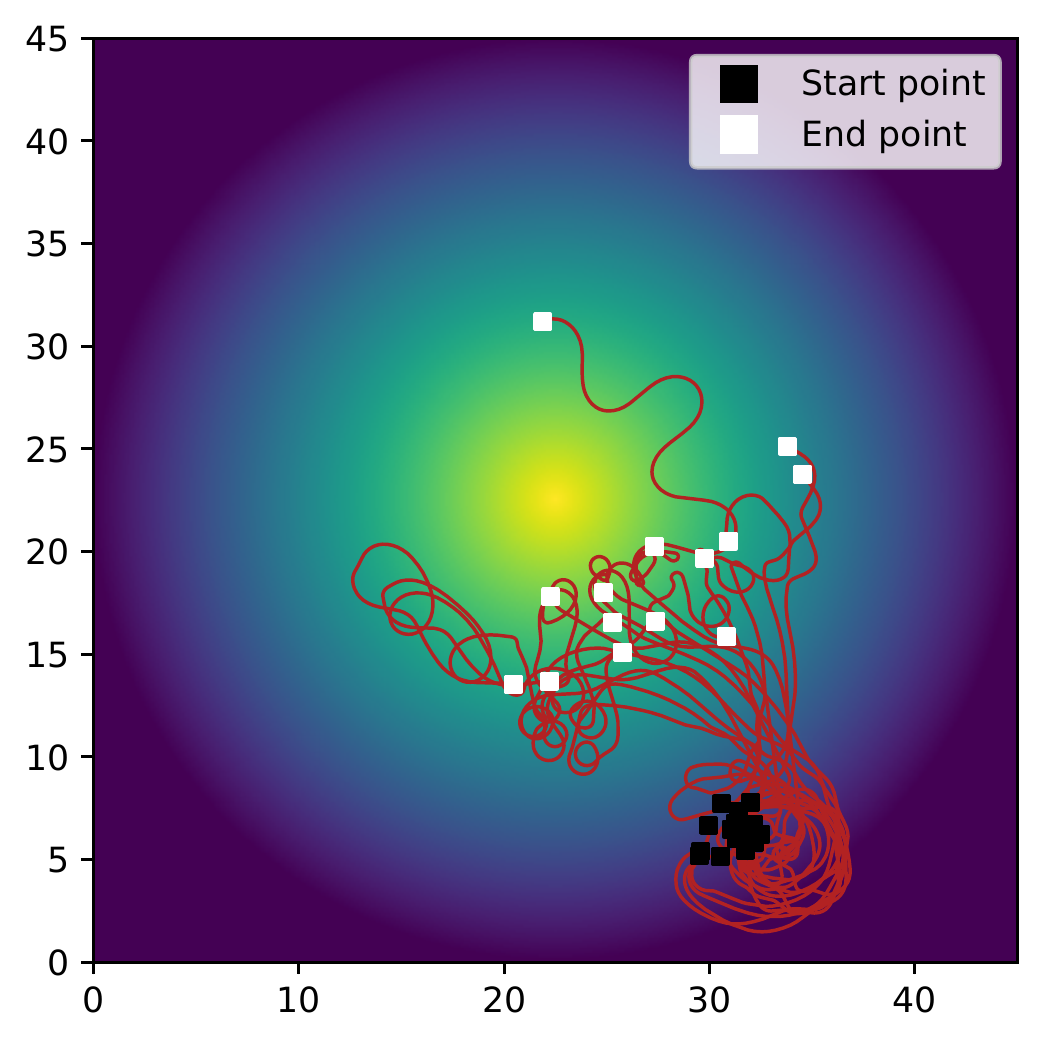}
         \caption{(d) 45x45}
         \label{fig:45x45}
     \end{subfigure}\vspace{-1em}
     \caption{No. of collisions and trajectories of the best controller per arena re-tested in the same environments. Please note the different scales in subplots b, c, d.}
     \label{fig:trajectories}     \vspace{-1.5em}
\end{figure}

\autoref{fig:trajectories}-a shows the the number of collisions over 600s timestamps in three arenas. We observe that the smaller arena size, the higher no. of collisions. Moreover, at the start of each run the swarm is very chaotic and collisions occur in all environments. Fortunately, it can be observed that the swarm resolves these collision itself. In addition, occasional group splits can also be observed. Since neither the ordered swarm behaviour is the main objective nor controllers are manually designed for it, we expect to see this splits time to time. What is interesting is that despite this splits, clusters continue to behave in the same way and eventually merge with the rest under the influence of the commonly sensed environmental feature.  \autoref{fig:trajectories}-(b,c,d) shows the trajectories of the best controller per arena re-tested in the same environments. The black blocks are the starting points of a swarm and the white blocks are the ending points. All of them are following the gradient well. They start with circling around, then moving towards the brighter area once they sense the different gradients. When they reach the brightest zone (yellow), they start circling again. A video of these behaviors can be found at \url{https://youtu.be/yhKFvpLa9iI}.

\section{Discussion}
The goal of this paper was to optimize the controllers of robots in a swarm for a task solely related to an environmental property and test whether a collective behavior emerges. Optimizing swarm behavior is often hard and requires vast knowledge on the specific task and the limitations of its members. Handwritten controllers therefore require a long process of prototyping which is sensitive to human biases, and limitation in engineered metrics for `good' collective behavior. In our setup we showed that even with a very simple task description, good performance can be reached with learned coordinated swarm behavior as an emergent property. 

From the results, the final best controller experiments showed that evolving in a bigger environment improves the controller in terms of flexibility. We suspect this to be caused by a less pronounced gradient in the bigger arena making the task more difficult. In other words, when evolved in a bigger arena a swarm needs to sample a bigger area to be able to sense the increasing gradient. This is because ratio between the swarm area and the total scalar field area is the smallest in the 45x45m environment. To cope with less salience of local values differences, the 45x45m flocks adopts a line formation strategy covering a big circle with more area (see \autoref{fig:order}-C1). This is slightly different than the spiral formation seen in \autoref{fig:order}-B1 employed by swarms evolved in the other arenas, and seem to maintain high fitness values when cross-validated. These behaviors can also be seen in the swarm trajectories (\autoref{fig:trajectories}). Lastly, we also see different behavior back in a less aligned rotation (low order) and more aligned forward motion (high order) with the best controller evolved in the largest environment. 

The alignment within the swarm members (\autoref{fig:order}) provides some interesting information on the swarm behavior. Here, we see a correlation between increasing order value and increasing mean scalar value seen by the swarm. Vice versa, lower order value correspond to a plateau/decrease in mean scalar value. This indicates that, when a swarm is becoming more aligned, the direction of movement is similar to the direction of increasing gradient. 

As a final remark, for our specific optimization we wanted the swarm to sense the increasing gradient (while a single one is not capable of doing it) and move towards the increasing local values. It should be noted that such a fitness could be any function that we want. This opens up the possibility to ignore standard engineering practices that enforce certain swarm behavior, optimize flocking rule parameters or design complex controllers --all of which may only be optimal for a specific task-- and continuously update our swarm as an adaptive system. The model-agnostic nature of a NN with an EA enables the emergence of different swarm strategies that we don't have to pre-define (think of different formations, sub-group sizes, search behavior, etc.). 

\section{Conclusions and Future Work}
We have successfully developed an evolutionary system for the automated design of controllers for robot swarms. As explained in the Introduction, this is a highly nontrivial feat because of the long and complex chain between genotypes and fitness values. To test our pipeline we evolved several swarms for the task of gradient following. Here, we found that the environmental property induced the emergence of collective behavior, a result that was not explicitly formalised in the objective function. To our best knowledge, there is only one other study that has achieved this \cite{witkowski2016emergence}, but we should note that it considered simulated point agents with signalling abilities which significantly simplified the task. Ongoing work is devoted to additional validation of our results using real robots. 

Interesting directions for future research include investigating more and different gradient landscapes in addition to the circular and unimodal one used in this study. Different gradient landscapes might be circular as well but with a multimodal structure or they can be in a linear form instead of circular. Moreover, instead of having a homogeneous swarm in terms of sensing capabilities, more investigation can be conducted where the sensing abilities of the individual robots differ (some may have more or better sensors than others, while using the same generic controller). This investigation can even be stretched by extending the evolutionary approach to swarms where the controllers are not necessarily identical as they are in this study.   

\begin{acks}
  This work is supported by Technology Innovation Institute, UAE.
\end{acks}

\bibliographystyle{ACM-Reference-Format}
\bibliography{acmart}

\end{document}